# STOCK2VEC: AN EMBEDDING TO IMPROVE PREDICTIVE MODELS FOR COMPANIES

ZIRUO YI

*Department of Computer Science & Engineering, University of North Texas*
*Denton, TX 76207, U.S.A.*
*E-mail: ZiruoYi@my.unt.edu*

TING XIAO[†], KAZ_ONYEAKAZI IJEOMA and RATNAM CHERAN

*Department of Information Science, University of North Texas*
*Denton, TX 76207, U.S.A.*
[†]*E-mail: Ting.Xiao@unt.edu*

YUVRAJ BAWEJA and PHILLIP NELSON

*Texas Academy of Math and Science*
*Denton, TX 76201, U.S.A.*

Building predictive models for companies often relies on inference using historical data of companies in the same industry sector. However, companies are similar across a variety of dimensions that should be leveraged in relevant prediction problems. This is particularly true for large, complex organizations which may not be well defined by a single industry and have no clear peers. To enable prediction using company information across a variety of dimensions, we create an embedding of company stocks, Stock2Vec, which can be easily added to any prediction model that applies to companies with associated stock prices. We describe the process of creating this rich vector representation from stock price fluctuations, and characterize what the dimensions represent. We then conduct comprehensive experiments to evaluate this embedding in applied machine learning problems in various business contexts. Our experiment results demonstrate that the four features in the Stock2Vec embedding can readily augment existing cross-company models and enhance cross-company predictions.

## 1. Introduction

Traditionally, predictive models are used to solve prediction problems of companies based on inference leveraging historical data of companies in the same industry. For instance, sales forecasting models can relate company sales to industry sales, and they also relate company variables to industry variables to solve short-, medium- and long-term prediction problems (Sharp *et al.*, 1982). However, many companies, especially those that are large and complex, are growing businesses across different industries so they do not simply belong to one specific industry. Moreover, some innovative companies do not have peers yet, so no existing industry can accurately define them. Therefore, only considering data from companies in the same industry is not enough and will lead to inaccurate predictions of companies. Besides industry, companies are similar to each other and share characteristics in other dimensions such as the geographic location and the company size. This raises a question: can we learn cross-company information in different dimensions to reduce the limitation of existing predictive models?

Companies are complex organizations and cannot be adequately represented by text-based embeddings due to three reasons: First, information about companies may not be fully available in a corpus. Second, information about companies may be outdated. Third, information about a company may need higher-order reasoning than what a text-based embedding can provide. Compared with texts, stock prices, which are the valuation of companies, are more informative. For instance, changes to the structure and function of a company can ultimately affect its stock price. Additionally, a stock reflects the cumulative prediction of many external observers to a company's performance. In essence, the fluctuations of stock prices are sensitive to a variety of company characteristics. Moreover, the factors that impact one company can also impact a similar company in many dimensions such as geography and industry. Therefore, creating a non-textual embedding based on stock price changes across companies over time is a good way to represent a company.

In this paper, we introduce Stock2Vec, an inexpensive but efficient embedding of company stocks to learn cross-company inference. Figure 1 illustrates the two-stage pipeline of our work. In the first stage, we use ordered daily price changes in S&P 500 stock over a 5-year span and the

Word2Vec algorithm to create an embedding for each stock, then we determine the most useful four features using machine learning models and obtain an optimized 4-dimensional embedding as our Stock2Vec embedding. In the second stage, we evaluate the Stock2Vec embedding by adding it to predictive models and working on two company characteristic prediction tasks: predicting the company's environmental impact and the company size with and without the Stock2Vec embedding. Experiments show that the addition of the Stock2Vec embedding enables a significant performance boost over using the original models only.

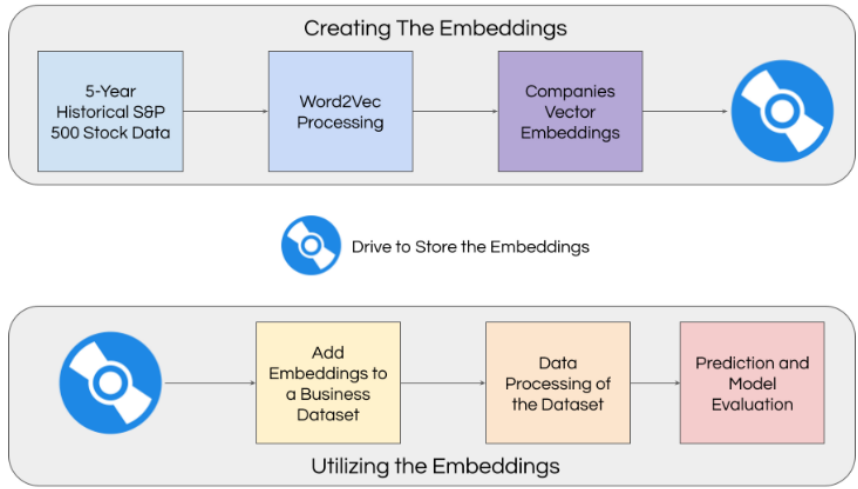

Fig. 1. Overview of the two-stage pipeline of creating and utilizing the Stock2Vec embedding.

Our contributions are summarized as follows: (i) We introduce Stock2Vec, an inexpensive but powerful embedding to represent companies. (ii) We present a detailed analysis of the Stock2Vec embedding. (iii) We achieve better performance on two company characteristic prediction tasks with the addition of Stock2Vec embedding to predictive models. We also make a discussion on the experiment results.

## 2. Literature Review

### 2.1. Stock predictions and machine learning

Stock prediction is important in data-driven decision making and deriving strategies. Stock forecasting has been focused on informative external data sources such as: accounting performance of companies, macroeconomic effects, etc. (Wang *et al.*, 2020). Machine learning methods can analyze large-scale financial datasets and outperform traditional statistical time series models. Therefore, Machine learning methods have been used along with historical stock prices in traditional technical analysis. Wu *et al.* (2019) propose that machine learning employs two major approaches in forecasting equity returns: The first method is Artificial Neural Network (ANN), which is used as a black box for standard factors correlated with equity returns. The second approach is forecasting the price time-series, in which price, trading volume, and other statistics are used to represent trading activities as a time-series. The researchers also identify three weaknesses in existing models used by computer scientists and financial professionals: undefined source of alpha, lack of heterogeneous sets used in models, and ineffectiveness in leveraging cross-sectional effects, which result in forecasting problems in the training pairs. Shen *et al.* (2012) postulate that features, which are used as inputs to algorithms, are mostly derived from industry-specific data. Since these approaches ignore pertinent information that may be available in other entities, the researchers break through the industry-centric barriers by building a trading model to utilize features that are not bound by the market.

## 2.2. Word embedding and Word2Vec algorithm

Knowledge representation is critical to learning inference and there are many ways to draw meaning from information. For instance, exploring similarities through comparison is used to measure similar states and circumstances that lead to predictive outcomes. One method to draw meaning through associative relationships between entities is embedding. Embeddings are low-dimensional translations of high-dimensional vectors, where vectors with stronger relationships are placed closer together. Word embeddings cluster similar words together and represent words in multi-dimensional vectors (Rizkallah *et al.*, 2020). Thus, word embeddings provide highly informative context about the relationship of words and transform seemingly arbitrary combinations of letters into representations that are useful for reasoning and generalization. Mikolov *et al.* (2013) find that word embeddings retain 70% of their original meaning and help models perform well in natural language processing tasks. In natural language processing, there are many applications for word embeddings, for example, being used in recommendation systems, named entity recognition, and sentiment analysis.

Researchers in Google introduced Word2Vec, a popular embedding that carries correlated features based on statistical relationships of words and yields a vector representation of words in sentences of large corpuses. The Word2Vec algorithm is a fusion of two learning models: Continuous Bag of Words (CBOW) and Skip-gram (Mikolov *et al.*, 2013). The Word2Vec embedding not only allows reasoning based on similarities but also preserves relative relationships among mathematical features. For instance, word embeddings can follow reasoning similarly to conceptual understandings such as "woman" + ("king" - "man") = "queen" and "swimming" + ("walked" – "walking") = "swam". The success of Word2Vec has led to a series of similar strategies that successfully create data representations in different forms and contexts. For example, Node2Vec (Grover and Leskovec, 2016) creates vector representations of nodes in a graph based on the context of connections within the graph. Doc2Vec (Lau and Baldwin, 2016) creates a vector representation of entire documents instead of words. And deep learning models such as XLNet are used to capture more complex and context-dependent relationships (Peters *et al.*, 2018; Yang *et al.*, 2019).

## 2.3. Stock2Vec works

The Word2Vec algorithm has been used as a sector embedding for predictions related to company stock attributes, and in some cases named as Stock2Vec. Stock2Vec in Wang *et al.* (2020)'s study learns the intrinsic relationships among stocks to make better predictions, in which more similar stocks are categorized closer to deploy interactions in the embedding. Lien Minh *et al.* (2018) also build a Stock2Vec embedding to predict stock movement. First, they extract and preprocess financial news from a financial news dataset, then label news and stock prices from the news as either negative or positive. Using labeled news and Harvard IV-4 dictionary, the Stock2Vec embedding is created and then trained by a Bidirectional gated recurrent unit (BGRU) on daily S&P 500 Index stock prices from Yahoo Finance. Finally, the trained Stock2Vec embedding is used to predict the effect of financial news on stock prices. The researchers also find that Stock2Vec is more effective than other embedding methods such as Glove and Word2Vec since it takes the sentiment value of words into consideration. Similarly, Lu *et al.* (2021) also train Stock2Vec on stock news and sentiment dictionaries. But they include political news and stock forum speech in sentiment analysis, and the Stock2Vec embedding is trained on CSI 300 stock data. They first use the Bayesian model for sentiment classification of the stock forum speech, then use Bidirectional Long Short Term Memory (Bi-LSTM) to extract trading data and investor sentiment index-related features, and finally use Contextual Long Short Term Memory (CLSTM) to integrate and process the political news within context. The stock trend is classified as positive with the probability of the stock going up and as negative if in the reverse case.

Compared with previous works, our work has two main differences: First, the Stock2Vec embeddings in existing works are derived with text data from news and sentiment dictionaries while our Stock2Vec embedding is generated on numeric data from daily price change of S&P 500 Index stock. This is premised on the fact that organizations within the same industry share similar growth or decline in Corporate Social Performance (CSP), which is asserted by Short *et al*. (2016). Second, previous works focus on predicting stock trends using company characteristics, but none of them leverage Word2Vec algorithm to build an embedding that utilizes stock prices to predict company characteristics. Dorfleitner *et al.* (2015) find that Environment, Social, Governance (ESG) scores play a major role in the decision-making process of managers and investors who are mindful of social responsibilities among other things. Crespi and Migliavacca (2020) underscore this by stating that the tendency of linear growth in the ESG score is enhanced by their size and profitability. Zumente and Lāce (2021) posit that ESG score has a correlation with the trading volume of company stock. Therefore, we use ESG rating and company size as our company characteristic prediction tasks, and we show that our Stock2Vec embedding can effectively predict company characteristics.

### 3. Stock2Vec Embedding

In this section, we first introduce the two datasets used and how to combine and preprocess them to the appropriate format for our work, then we describe the detailed steps to generate our Stock2Vec embedding.

#### *3.1. Dataset and preprocessing*

The dataset used in our work is a combination of two datasets. The first dataset is the S&P 500 stock data[1], which is about historical stock prices for all companies in the S&P 500 index from 2013 to 2018. This dataset contains daily trading information including trading date, opening price, highest price, lowest price, closing price, the number of shares traded, and company stock name. The other dataset is the S&P 500 companies with Financial Information[2], from which we can get S&P 500 companies' names, sectors, and symbols.

Our data preprocessing step is done using Python. We first combine two datasets according to the companies' stock names and then remove irrelevant information such as the number of shares traded. After the cleanup, the preprocessed dataset contains the daily price change of the S&P 500 stocks, corresponding sectors, and dates. Each of the 505 stocks is represented by 1826 data points (365*5+1) containing the daily stock price change for each day from 2013 to 2018.

#### *3.2. Creating the Stock2Vec embedding*

We first order the stocks in the dataset based on their daily price change for each day. The purpose of this step is to create a 'sentence', which is a sequence of stock fluctuation information for each day. The daily stock price change is calculated as the difference between the closing and opening price for one day, and then divided by the opening price.

The 1826 generated stock sentences are then input to the Word2Vec algorithm which results in a high-dimensional Word2Vec embedding. One problem is that what each feature represents is unclear, because feature values are arbitrary numbers that describe relationships between objects. As a result, the optimal number of features required to represent a company is unclear. To decide an appropriate dimensionality for our Stock2Vec embedding, one efficient way is to observe the utilization of the embedding in prediction problems and identify the minimum number of dimensions necessary to achieve a high accuracy. Specifically, we define an industry sector prediction task based on the daily stock price changes, then split the data into a 70% training set and a 30% testing set respectively. Next, we use the Word2Vec embedding on four classifiers: Gaussian Naive Bayes (Rish *et al.*, 2001), Support Vector Machines (SVM) (Zhang, 2012), Decision Tree (Safavian and Landgrebe, 1991), and the Random Forest (Breiman, 2001). If one classifier gets a higher accuracy with the Word2Vec embedding at a given dimensionality, it stands to reason that the Word2Vec embedding with that dimensionality can better represent the original data. Afterwards, we train the Word2Vec embedding on the training set and then classify each company to its corresponding sector with the remaining test set.

---

[1] https://www.kaggle.com/camnugent/sandp500

[2] https://datahub.io/core/s-and-p-500-companies-financials#resource-s-and-p-500-companies-financials_zip

Figure 2 demonstrates the accuracy of each classifier's prediction based on the Word2Vec embedding with varying dimensions. The accuracy of four classifiers begin to plateau at roughly 4 dimensions, which means that afterwards increasing dimensions of embedding only increases the prediction accuracy by a negligible amount. Thus, we decide the optimal dimensionality of the embedding is four and named the new four-dimensional embedding Stock2Vec. Obviously, Stock2Vec is a lower dimensional embedding for a simpler but more meaningful representation of each company.

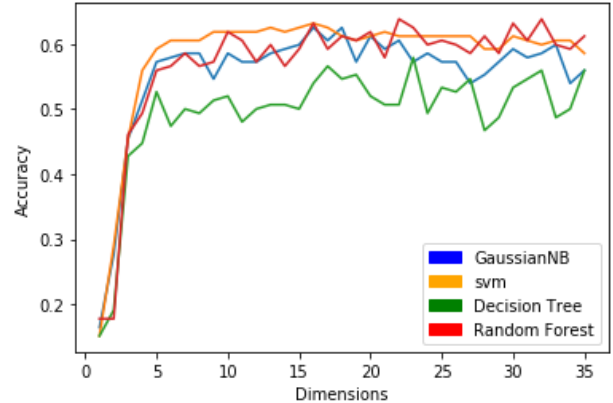

Fig. 2. Sector prediction accuracy of four classifiers based on the Word2Vec embedding with varying dimensions. It can be noted that the accuracy of all classifiers plateau when the embedding has more than four dimensions.

## 4. Stock2Vec Embedding Analysis

We analyze the nature of our Stock2Vec embedding from two angles. We first check whether the Stock2Vec embedding is optimal, then we explore whether the Stock2Vec embedding leads to prediction mistakes.

### 4.1. Checking the Stock2Vec embedding

To further confirm that our four-dimensional Stock2Vec embedding is optimal, we apply Principal Component Analysis (PCA) (Shlens, 2014) to create a graphical representation of the Word2Vec embedding and visualize it on a cartesian coordinate plane. PCA is an algorithm utilized to compress data into a lower-dimensional space but is less adept at retaining relationships between other objects. If the Word2Vec embedding gets a larger explained variance at a given dimensionality, that means it represents a higher percentage of information of the original data. Figure 3 below shows that when the dimensionality of embedding is four, the embedding captures over 95% of the variance in the dataset, which means that the four-dimensional embedding maintains around 95% information of the original data. Moreover, the Word2Vec embeddings with less than four dimensions maintain much less original information while embeddings with more than four dimensions do not maintain obviously more original information. Although embeddings with more than four dimensions are a little bit more informative, higher-dimensional embedding requires more computing resources, which is very costly. Therefore, four is confirmed as the most appropriate dimensionality and is applied to our Stock2Vec embedding.

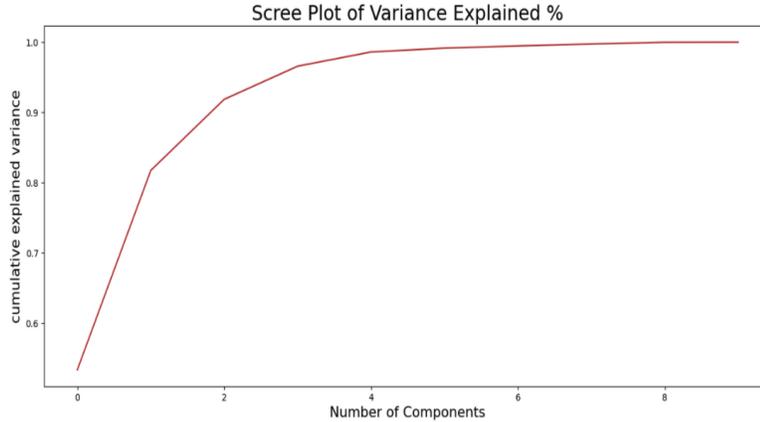

Fig. 3. Principal Component Analysis of the Wor2Vec embedding in varying dimensions and their corresponding explained variances. It shows how well the Stock2Vec embedding can capture the information in the original data.

*4.2. Characterizing the Stock2Vec embedding*

As shown in Figure 2, the SVM model provides a more stable metric of the accuracy out of the four classifiers. Thus, we use the SVM model to discuss the company sector prediction results. The SVM classifier is able to consistently attain around 60% accuracy when predicting sectors utilizing our Stock2Vec embedding, which indicates that about 60% of the information of a given company could be represented by the Stock2Vec embedding. There are mistakes in the prediction results and the SVM classifier determines which sectors are commonly confused and therefore are more closely related. To analyze the prediction mistakes, we create a confusion matrix to visualize these confusions.

Figure 4 shows the confusion matrix produced by the SVM classifier, in which the commonly mistaken sectors can be clearly identified. The sectors along the vertical axis of the matrix indicate the companies' actual sectors while the sectors along the horizontal axis indicate the sectors predicted by the SVM classifier. Therefore, accurate predictions lie along the diagonal and confusions lie outside the diagonal. Sectors such as Health Care and Information Technology are often misidentified, but these sectors have overlap in their services. This fact suggests that most of the incorrect predictions are due to the similarities between sectors, which means that these prediction mistakes are not caused by our Stock2Vec embedding.

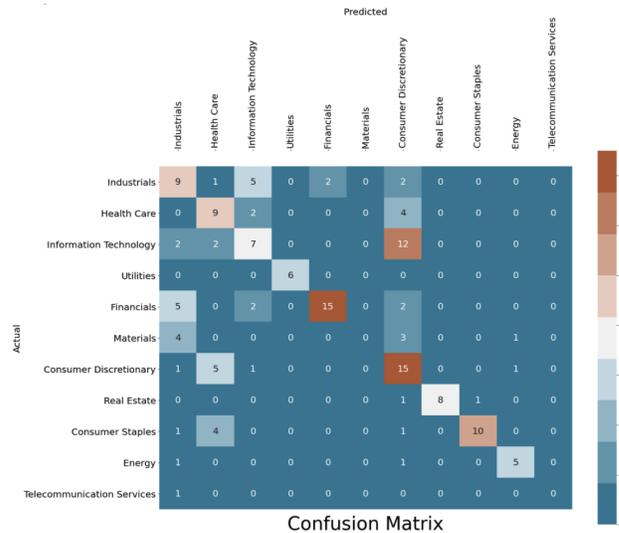

Fig. 4. Confusion matrix of the SVM classifier predicting companies' sectors using the Stock2Vec embedding. Redder areas indicate a greater amount of confusion while bluer areas indicate little to no confusion.

This conclusion is further supported by Figure 5, which is the PCA plot showing the clusters formed by the companies' sectors. Sectors such as Utilities and Energy have few errors in the confusion matrix, and the clusters formed by them on the plot are far from clusters formed by other sectors. This fact suggests that Utilities and Energy do not share many traits with other sectors. In contrast, the clusters formed by highly confused sectors such as Industrials, show a wider distribution that overlaps with various clusters formed by other sectors. Therefore, our Stock2Vec embedding makes it easy to visualize and interpret the relationships between companies and their corresponding sectors.

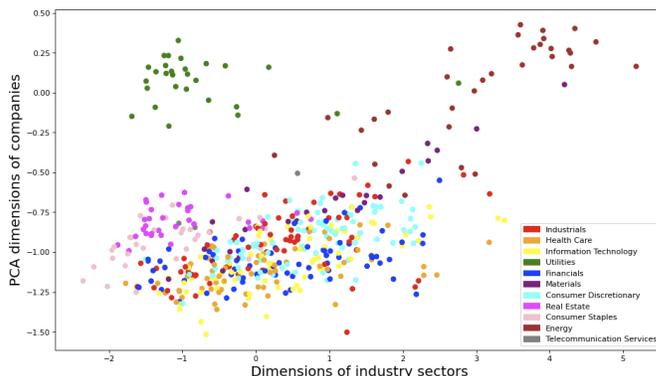

Fig. 5. PCA plot of colored-coded companies and their corresponding sectors. The corresponding relationships are indicated down right.

## 5. Applications

To evaluate the viability of the Stock2Vec embedding, we design two prediction tasks of company characteristics: predicting the environmental impact of companies and predicting the size of companies. We also compare the results of using and not using the Stock2Vec embedding in these two tasks.

### 5.1. Data

For the prediction task of companies' environmental impact, we gather the Environmental, Social, and Governance (ESG) ratings' data from Sustainalytics web portal[3] and built an ESG ratings' dataset. For the prediction task of companies' size, we obtain an employee-count dataset from Liberated Stock Trader website[4]. We split our datasets into a 70% training set and a 30% testing set.

### 5.2. Experiments and results

For the prediction task of companies' environmental impact, we first perform a linear regression model using the ESG ratings dataset. Then we add the four features in the Stock2Vec embedding to the ESG ratings dataset from four dimensions (Dim1, Dim2, Dim3, and Dim4), and run the linear regression model again on the enlarged dataset. The same steps are repeated on the prediction task of companies' size. And for both tasks, we use the Ordinary Least Squares (OLS) estimator, which is a linear least squares method to estimate the unknown parameters in a linear regression model, to fit the model to the datasets.

As seen in Figure 6, when the four additional features are not added to the two datasets, the linear regression model achieves $R^2$ values of 10.5% for the prediction of employee count, and 2.0% for the prediction of ESG rating. After adding our Stock2Vec embedding, the $R^2$ value of the prediction of employee count increases from 10.5% to 12.4%, noting a 1.9% improvement. Similarly, the $R^2$ value for the prediction of ESG rating rises from 2.0% to 11.3%, which is a 9.3%

---

boost. Therefore, the linear regression model receives a drastic improvement in accuracy on the two prediction tasks with the help of our Stock2Vec embedding.

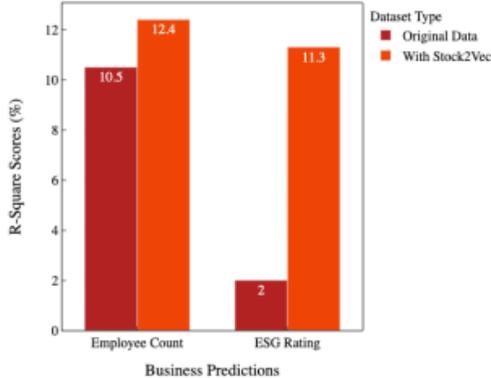

Fig. 6. $R^2$ values achieved by the regression models for the ESG rating prediction and the employee count prediction. For both tasks, the $R^2$ values increase when the Stock2Vec embedding is included.

To confirm the importance of the Sotck2Vec embedding, we need to interpret the model and determine which features are most beneficial for the prediction tasks. Feature importance is the most useful interpretation tool to identify important features, so we use the Random Forest Importance (Strobl, 2008), which is a reliable and efficient technique and can be applied to any model, to extrapolate the significance of each feature in the OLS model for the ESG Ratings prediction task and the employee-count prediction task.

As shown in Figure 7, the importance score indicates the significance of each feature in the prediction making process, and the four features in our Stock2Vec embedding (Dim1-4 represent four dimensions that the four features add to) contribute the most to the ESG Ratings OLS model. Similarly in Figure 8, the Stock2Vec features also bear high importance compared with other features used in the employee-count prediction. The feature importance scores confirm that the Stock2Vec embedding plays a major role in raising the R2 values of the prediction tasks.

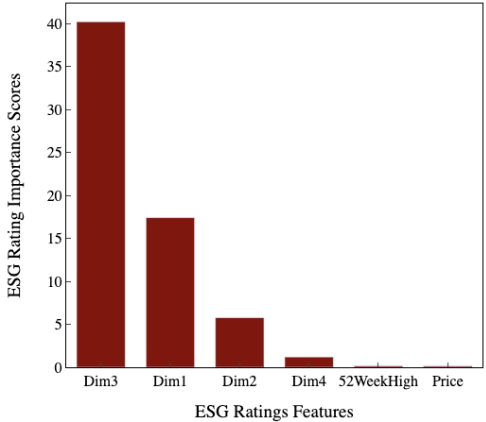

Fig. 7. Random Forest Importance score of each feature used in the ESG Ratings OLS model.

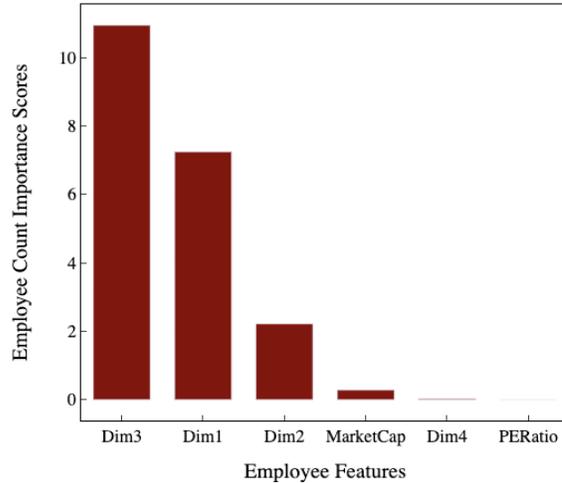
Fig. 8. Random Forest Importance score of each feature used in the employee-count OLS model.

### 5.3. Discussion

#### 5.3.1. Employee-count OLS model

We use the residuals plot to analyze the variance of error in the employee-count OLS regressor. As shown in Figure 9, the residual points are randomly dispersed around the horizontal axis, which indicates that the model is appropriate for the data.

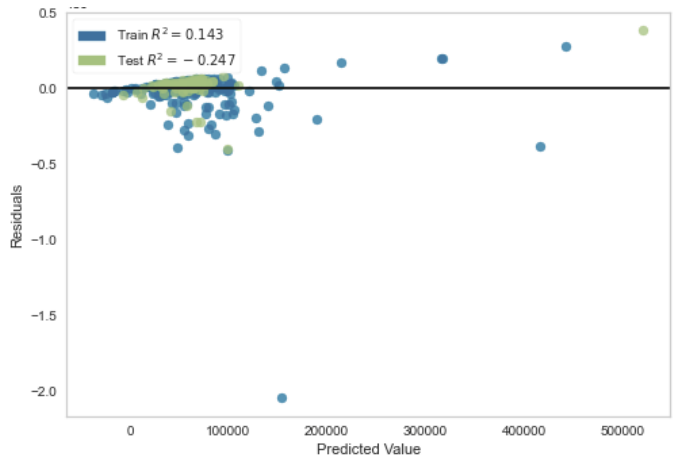
Fig. 9. Residuals for the employee-count OLS model

Figure 10 shows that the R-squared, which measures the accuracy of the model, is 10.5% for the employee-count model without adding the four features in the Stock2Vec embedding. And after adding the Stock2Vec features, as shown in Figure 11, the R-squared is 12.4% for the model. Additionally, the p-values of the four Stock2Vec features X3, X4, X5, X6 are lower than the p-value of the variable X2, which is 0.919. X2 is the MarketCap feature from the original employee-count dataset. Hence, the addition of the four features in the Stock2Vec embedding helps improve the model.

```
==============================================================================
Dep. Variable:      NumberOfEmployees   R-squared:                       0.105
Model:                            OLS   Adj. R-squared:                  0.100
Method:                 Least Squares   F-statistic:                     21.21
Date:                Tue, 02 Feb 2021   Prob (F-statistic):           1.94e-09
Time:                        16:15:24   Log-Likelihood:                -4846.9
No. Observations:                 366   AIC:                             9700.
Df Residuals:                     363   BIC:                             9712.
Df Model:                           2
Covariance Type:            nonrobust
==============================================================================
                 coef    std err          t      P>|t|      [0.025      0.975]
------------------------------------------------------------------------------
const        4.272e+04   8254.598      5.175      0.000    2.65e+04    5.9e+04
x1            245.4177     37.685      6.512      0.000     171.309     319.527
x2              0.7379     64.785      0.011      0.991    -126.664     128.139
==============================================================================
Omnibus:                      629.039   Durbin-Watson:                   0.581
Prob(Omnibus):                  0.000   Jarque-Bera (JB):           303098.303
Skew:                           9.783   Prob(JB):                         0.00
Kurtosis:                     142.615   Cond. No.                         238.
==============================================================================
```

Fig. 10. Prediction summary of the employee-count OLS model without the Stock2Vec embedding.

```
==============================================================================
Dep. Variable:      NumberOfEmployees   R-squared:                       0.124
Model:                            OLS   Adj. R-squared:                  0.109
Method:                 Least Squares   F-statistic:                     8.447
Date:                Thu, 04 Feb 2021   Prob (F-statistic):           1.38e-08
Time:                        07:14:52   Log-Likelihood:                -4843.0
No. Observations:                 366   AIC:                             9700.
Df Residuals:                     359   BIC:                             9727.
Df Model:                           6
Covariance Type:            nonrobust
==============================================================================
                 coef    std err          t      P>|t|      [0.025      0.975]
------------------------------------------------------------------------------
const        2.943e+05   2.67e+05      1.101      0.271   -2.31e+05    8.2e+05
x1            237.0795     38.374      6.178      0.000     161.613     312.546
x2             -6.5753     64.656     -0.102      0.919    -133.728     120.577
x3          -1.631e+05   1.65e+05     -0.988      0.324   -4.88e+05    1.62e+05
x4           8.814e+04   9.55e+04      0.923      0.357   -9.96e+04    2.76e+05
x5           1.822e+05   1.94e+05      0.938      0.349       -2e+05    5.64e+05
x6           4043.7607   1.89e+04      0.214      0.831   -3.32e+04    4.13e+04
==============================================================================
Omnibus:                      633.299   Durbin-Watson:                   0.598
Prob(Omnibus):                  0.000   Jarque-Bera (JB):           315535.382
Skew:                           9.908   Prob(JB):                         0.00
Kurtosis:                     145.472   Cond. No.                      1.11e+04
==============================================================================
```

Fig. 11. Prediction summary of the employee-count OLS model with the Stock2Vec embedding.

### 5.3.2. ESG ratings OLS model

Figure 12 shows the residuals plot of the ESG ratings OLS model, which is used to analyze the variance of error in the ESG ratings OLS regressor. We find that the residual points are randomly dispersed around the horizontal axis of the predicted value, which confirms that it is a well fitted model.

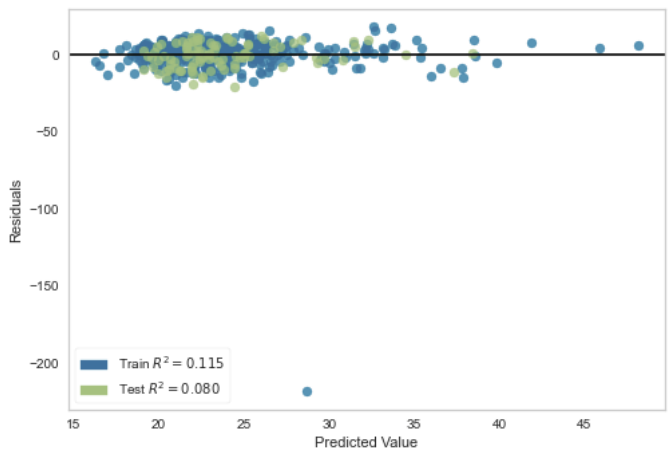

Fig. 12. Residuals for the ESG ratings OLS model

Figure 13 and Figure 14 show that the R-squared of the ESG ratings model performs better when we add the four features in the Stock2Vec embedding. Initially in Figure 13, the accuracy score is only 2.0%. After adding the four features, the accuracy immediately boosts to 11.3 % in Figure 14. Consequently, the four Stock2Vec features represented by X8, X9, X10 and X11 are statistically significant with p-value scores of 0.003, 0.035, 0.008, and 0.001 respectively. All the p-values of the four features are less than the p-value of X2, which is 0.980. This indicates that the four features have a significant impact on the ESG ratings prediction. In conclusion, the addition of the Stock2Vec embedding significantly improves the model.

```
==============================================================================
Dep. Variable:              ESGRating   R-squared:                       0.020
Model:                            OLS   Adj. R-squared:                  0.005
Method:                 Least Squares   F-statistic:                     1.317
Date:                Tue, 02 Feb 2021   Prob (F-statistic):              0.240
Time:                        13:52:46   Log-Likelihood:                -1833.9
No. Observations:                 461   AIC:                             3684.
Df Residuals:                     453   BIC:                             3717.
Df Model:                           7
Covariance Type:            nonrobust
==============================================================================
                 coef    std err          t      P>|t|      [0.025      0.975]
------------------------------------------------------------------------------
const         24.1842      1.387     17.435      0.000      21.458      26.910
x1            -0.0175      0.047     -0.375      0.708      -0.109       0.074
x2            -0.0036      0.015     -0.235      0.815      -0.033       0.026
x3            -0.0717      0.456     -0.157      0.875      -0.968       0.824
x4            -0.2598      0.142     -1.835      0.067      -0.538       0.018
x5            -0.0241      0.045     -0.538      0.591      -0.112       0.064
x6             0.0601      0.044      1.377      0.169      -0.026       0.146
x7          1.449e-11   7.73e-12      1.875      0.061   -6.97e-13    2.97e-11
==============================================================================
Omnibus:                      821.756   Durbin-Watson:                   1.969
Prob(Omnibus):                  0.000   Jarque-Bera (JB):           638692.856
Skew:                          10.850   Prob(JB):                         0.00
Kurtosis:                     184.052   Cond. No.                     2.36e+11
==============================================================================
```

Fig. 13. Prediction summary of the ESG ratings OLS model without the Stock2Vec embedding.

```
==============================================================================
Dep. Variable:              ESGRating   R-squared:                       0.113
Model:                            OLS   Adj. R-squared:                  0.091
Method:                 Least Squares   F-statistic:                     5.193
Date:                Thu, 04 Feb 2021   Prob (F-statistic):           1.04e-07
Time:                        08:05:21   Log-Likelihood:                -1810.9
No. Observations:                 461   AIC:                             3646.
Df Residuals:                     449   BIC:                             3695.
Df Model:                          11
Covariance Type:            nonrobust
==============================================================================
                 coef    std err          t      P>|t|      [0.025      0.975]
------------------------------------------------------------------------------
const        -24.3845     18.063     -1.350      0.178     -59.884      11.115
x1            -0.0184      0.046     -0.396      0.692      -0.109       0.073
x2            -0.0004      0.015     -0.025      0.980      -0.029       0.028
x3            -0.7662      0.509     -1.506      0.133      -1.766       0.233
x4            -0.0406      0.140     -0.289      0.772      -0.317       0.235
x5            -0.0095      0.047     -0.204      0.839      -0.101       0.082
x6             0.0373      0.044      0.842      0.400      -0.050       0.124
x7          1.684e-11   7.56e-12      2.228      0.026    1.99e-12    3.17e-11
x8            32.8607     10.930      3.007      0.003      11.381      54.340
x9           -13.7810      6.504     -2.119      0.035     -26.563      -0.999
x10          -34.5832     12.999     -2.660      0.008     -60.130      -9.037
x11           -4.4503      1.368     -3.253      0.001      -7.139      -1.762
==============================================================================
Omnibus:                      870.489   Durbin-Watson:                   2.001
Prob(Omnibus):                  0.000   Jarque-Bera (JB):           881909.869
Skew:                          12.192   Prob(JB):                         0.00
Kurtosis:                     215.881   Cond. No.                     4.44e+12
==============================================================================
```

Fig. 14. Prediction summary of the ESG ratings OLS model with the Stock2Vec embedding.

## 6. Conclusion

In this paper, we present Stock2Vec, an efficient embedding to represent companies. We use two datasets and two company characteristic prediction tasks to analyze and evaluate the capability of the Stock2Vec embedding. The experiment results demonstrate that the Stock2Vec embedding can characterize companies using only their stock information, and that company characteristic prediction results can be improved by adding the four Stock2Vec features to predictive models. Moreover, the construction of the Stock2Vec embedding is fast and can be shared readily, which is less expensive than other existing methods. Our Stock2Vec embedding breaks the limitation of existing predictive models and provides insights to future studies of business such as stock market analysis and predictive models generation.

**References**

Breiman, L. Random Forests. Machine Learning 45, 5–32 (2001). https://doi.org/10.1023/A:1010933404324

Crespi, F., & Migliavacca, M. (2020). The Determinants of ESG Rating in the Financial Industry: The Same Old Story or a Different Tale? Sustainability: Science Practice and Policy, 12(16), 6398.

Dorfleitner, G., Halbritter, G., & Nguyen, M. (2015). Measuring the level and risk of corporate responsibility – An empirical comparison of different ESG rating approaches. In Journal of Asset Management (Vol. 16, Issue 7, pp. 450–466). https://doi.org/10.1057/jam.2015.31

Grover, A., & Leskovec, J. (2016). node2vec: Scalable Feature Learning for Networks. Proceedings of the 22nd ACM SIGKDD International Conference on Knowledge Discovery and Data Mining, 855–864.

J.A. Sharp, D.H.R. Price (1982). Industry modelling, Omega, Volume 10, Issue 3, Pages 237-247, ISSN 0305-0483

Lau, J. H., & Baldwin, T. (2016). An Empirical Evaluation of doc2vec with Practical Insights into Document Embedding Generation. In Proceedings of the 1st Workshop on Representation Learning for NLP. https://doi.org/10.18653/v1/w16-1609

Lien Minh, D., Sadeghi-Niaraki, A., Huy, H. D., Min, K., & Moon, H. (2018). Deep Learning Approach for Short-Term Stock Trends Prediction Based on Two-Stream Gated Recurrent Unit Network. IEEE Access, 6, 55392–55404.

Lu, R., & Lu, M. (2021). Stock Trend Prediction Algorithm Based on Deep Recurrent Neural Network. Wireless Communications and Mobile Computing, 2021.

Mikolov, T., Chen, K., Corrado, G., & Dean, J. (2013). Efficient Estimation of Word Representations in Vector Space. In arXiv [cs.CL]. arXiv. http://arxiv.org/abs/1301.3781

Mikolov, T., Sutskever, I., Chen, K., Corrado, G. S., & Dean, J. (2013). Distributed representations of words and phrases and their compositionality. In Advances in neural information processing systems (pp. 3111-3119).

Peters, M. E., Neumann, M., Iyyer, M., Gardner, M., Clark, C., Lee, K., & Zettlemoyer, L. (2018). Deep contextualized word representations. Proceedings of the 2018 Conference of the North American Chapter of the Association for Computational Linguistics: Human Language Technologies, Volume 1 (Long Papers).

Rish, I., & Others. (2001). An empirical study of the naive Bayes classifier. IJCAI 2001 Workshop on Empirical Methods in Artificial Intelligence, 3, 41–46.

Rizkallah, S., Atiya, A. F., & Shaheen, S. (2020). A Polarity Capturing Sphere for Word to Vector Representation. NATO Advanced Science Institutes Series E: Applied Sciences, 10(12), 4386.

Safavian, S. R., & Landgrebe, D. (1991). A survey of decision tree classifier methodology. IEEE Transactions on Systems, Man, and Cybernetics, 21(3), 660–674.

Shen, S., Jiang, H., & Zhang, T. (2012). Stock market forecasting using machine learning algorithms. Department of Electrical Engineering, Stanford University, Stanford, CA, 1-5.

Shlens, J. (2014). A tutorial on principal component analysis. arXiv preprint arXiv:1404.1100.

Short, J. C., McKenny, A. F., Ketchen, D. J., Snow, C. C., & Hult, G. T. M. (2016). An Empirical Examination of Firm, Industry, and Temporal Effects on Corporate Social Performance. Business & Society, 55(8), 1122–1156.

Strobl, C., & Zeileis, A. (2008). Danger: High power! – exploring the statistical properties of a test for random forest variable importance

Wang, X., Wang, Y., Weng, B., & Vinel, A. (2020). Stock2Vec: A Hybrid Deep Learning Framework for Stock Market Prediction with Representation Learning and Temporal Convolutional Network. In arXiv [q-fin.ST]. arXiv. http://arxiv.org/abs/2010.01197

Wu, Q., Brinton, C. G., Zhang, Z., Pizzoferrato, A., Liu, Z., & Cucuringu, M. (2019). Equity2Vec: End-to-end deep learning framework for cross-sectional asset pricing. In arXiv [cs.LG]. arXiv. http://arxiv.org/abs/1909.04497

Yang, Z., Dai, Z., Yang, Y., Carbonell, J., Salakhutdinov, R. R., & Le, Q. V. (2019). Xlnet: Generalized autoregressive pretraining for language understanding. Advances in neural information processing systems, 32.

Zhang, Y. (2012). Support Vector Machine Classification Algorithm and Its Application. Information Computing and Applications, 179–186.

Zumente, I., & Lāce, N. (2021). ESG Rating—Necessity for the Investor or the Company? Sustainability: Science Practice and Policy, 13(16), 8940.